\documentclass{article}

\usepackage{PRIMEarxiv}

\usepackage[utf8]{inputenc} % allow utf-8 input
\usepackage[T1]{fontenc}    % use 8-bit T1 fonts
\usepackage{hyperref}       % hyperlinks
\usepackage{url}            % simple URL typesetting
\usepackage{booktabs}       % professional-quality tables
\usepackage{amsfonts}       % blackboard math symbols
\usepackage{nicefrac}       % compact symbols for 1/2, etc.
\usepackage{microtype}      % microtypography
\usepackage{lipsum}
\usepackage{fancyhdr}       % header
\usepackage{graphicx}       % graphics
\graphicspath{{media/}}     % organize your images and other figures under media/ folder
\usepackage{enumitem}
\usepackage{multirow}
\usepackage{caption}
\usepackage{subcaption}
\usepackage{times}
\usepackage{epsfig}
\usepackage{graphicx}
\usepackage{amsmath}
\usepackage{amssymb}
\usepackage{pbox}

\newcommand{\loss}{L}

\title{NVAutoNet: Fast and Accurate 360$^{\circ}$ 3D Visual Perception For Self Driving
}

\author{
\small{Trung Pham\thanks{Corresponding author: Trung Pham (trungp@nvidia.com)} , Mehran Maghoumi, Wanli Jiang, \break Bala Siva Sashank Jujjavarapu, Mehdi Sajjadi,}\\
\small{Xin Liu, Hsuan-Chu Lin, Bor-Jeng Chen, Giang Truong, Chao Fang, Junghyun Kwon, Minwoo Park}\\
NVIDIA \\
}

\begin{document}
\maketitle

\begin{figure}[h!]
\centering
\includegraphics[width=0.99\textwidth]{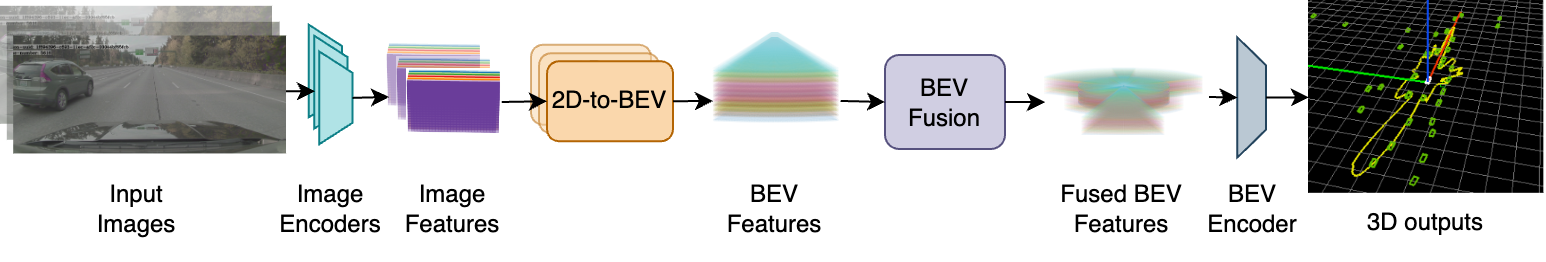}
\caption{NVAutoNet overview. Surround view images are input to CNN backbones to extract 2D features, which are uplifted and fused into a unified BEV feature map. The generated BEV features are then encoded by a BEV backbone. Finally 3D signals are predicted by various 3D perception heads.}
\label{fig:nvautonet}  
\end{figure}
\begin{table}[h!]
	\begin{tabular}{lllll}
	\hline
	Inputs & Outputs & Latency & Detection Range & 	Training data \\
	8 cameras & 3D signals  & 18ms (53fps) & 200 meters & 2.2M scenes\\
	\hline \hline 
    Architecture & Fusion & Modular design & End-to-end training & In-car tested \\ 
	 CNN + MLP & Mid-level & \checkmark  & \checkmark  & \checkmark  \\
	\hline
	\end{tabular}
	\centering
	\caption{NVAutoNet highlights.}
	\label{tab:nvautonet_highlights}
\end{table}

%%%%%%%%% ABSTRACT
\begin{abstract}
Achieving robust and real-time 3D perception is fundamental for autonomous vehicles. While most existing 3D perception methods prioritize detection accuracy, they often overlook critical aspects such as computational efficiency, onboard chip deployment friendliness, resilience to sensor mounting deviations, and adaptability to various vehicle types. To address these challenges, we present NVAutoNet: a specialized Bird's-Eye-View (BEV) perception network tailored explicitly for automated vehicles. NVAutoNet takes synchronized camera images as input and predicts 3D signals like obstacles, freespaces, and parking spaces. The core of NVAutoNet's architecture (image and BEV backbones) relies on efficient convolutional networks, optimized for high performance using TensorRT. More importantly, our image-to-BEV transformation employs simple linear layers and BEV look-up tables, ensuring rapid inference speed. Trained on an extensive proprietary dataset, NVAutoNet consistently achieves elevated perception accuracy, operating remarkably at 53 frames per second on the NVIDIA DRIVE Orin SoC. Notably, NVAutoNet demonstrates resilience to sensor mounting deviations arising from diverse car models. Moreover, NVAutoNet excels in adapting to varied vehicle types, facilitated by inexpensive model fine-tuning procedures that expedite compatibility adjustments.
\end{abstract}

%%%%%%%%% BODY TEXT
\section{Introduction}
\label{sec:intro}
Autonomous vehicles must accurately understand their 3D surroundings, but achieving this by camera sensors only is complex. In early days, individual monocular camera perception modules are run separately, and their outputs are then merged to create a unified 3D picture. However, this approach faces challenges. Errors from independent monocular modules, like under or overestimation, vary and their error models are often unknown. Consequently, merging these results becomes tricky, often causing false positives. Nonetheless, maintaining camera independence enhances product safety by reducing shared failure risks.

Another vital consideration is production scalability. Car manufacturers typically produce various car models like SUVs, sedans, sport cars, and trucks, each with distinct sizes and camera positions. Thus, a camera perception system that can handle diverse camera angles, positions, radial distortions, and focal lengths becomes essential for scalability. Equally important, the system must operate in real-time, within a low-powered, shared compute budget on a System on Chip (SoC), especially since other independent radar or ultrasonic modules might run concurrently.

\textbf{Design principles:} We aim to design a perception network with the following principles:
\begin{itemize}[noitemsep,nolistsep]
\item \textbf{Precise 3D Perception}: The projected 3D predictions seamlessly align with the content of 2D image views.
\item \textbf{Extended Range Detection}: The network excels in detecting objects at considerable distances, reaching up to 200 meters.
\item \textbf{Efficiency at Its Core}: Operating in real-time on edge devices, such as the NVIDIA DRIVE Orin SoC, the network ensures seamless responsiveness.
\item \textbf{Resilience to Camera Variability}: The network maintains robust functionality even in the presence of camera dropouts.
\item \textbf{Holistic Machine Learning}: Requiring no post-processing, the network's performance scales effectively with the volume of data.
\item \textbf{Adaptive Scalability}: Straightforward customization for diverse vehicle platforms underscores the network's scalability.
\end{itemize}

\textbf{Key components and features:} Guided by the aforementioned design principles, we introduce NVAutoNet, a camera perception network featuring the following essential components and attributes. (1) CNN based image feature extractors are meticulously tailored using hardware-aware neural architecture search (NAS) to achieve both high accuracy and low latency. (2) Multi-camera fusion occurs at the BEV level, combining the strengths of early and late fusion methods. (3) Perspective-to-BEV view transformation is efficiently executed through column-wise MultiLayer Perceptron (MLP) layers and BEV look-up tables. (4) All perception tasks, including freespace perception, are structured as set prediction tasks, streamlining processes and negating the need for resource-intensive post-processing like clustering, boundary extraction, and curve fitting. (5) A simple loss balancing algorithm greatly facilitates the process of multi-task learning. Figure \ref{fig:nvautonet} shows an overview of NVAutoNet and Table \ref{tab:nvautonet_highlights} summarizes its key features.

%The remainder of the paper is organized as follows. Section 2 discusses related work. Sections 3 explains NVAutoNet with details. Sections 4 and 5 explains NVAutoNet with details. Section 4 provides extensive experimental results. Lastly, section 5 concludes the paper and discusses a couple of future directions for autonomous driving (AV) perception.

\section{Related Work}
%To the best of our knowledge, there was no previously published work that has achieved all the goals listed above. In this section, we summarize methods which are related to one or more components of our work.
\textbf{Perception for Self-Driving.} In the realm of self-driving, the perception module's responsibility encompasses the detection and identification of all static and moving objects in the surrounding environment. This roster includes obstacles, road markings, lanes, road boundaries, traffic lights, traffic signs, parking spaces, free spaces, and more. Existing research has predominantly focused on the 3D object detection task, with notable methodologies like DETR3D \cite{Wang2021}, PETR \cite{LiuWZS22}, BEVFormer \cite{bevformer}, BEVDet \cite{huang2021bevdet}, and BEVDepth \cite{Li2022BEVDepthAO}. In contrast, our NVAutoNet network tackles multiple tasks concurrently, including 3D obstacles, 3D freespaces, and 3D parking spaces. Modern 3D object detection methods have embraced set prediction approaches to obviate the need for NMS post-processing, a trend that NVAutoNet seamlessly follows. Recently, the domain of online mapping perception \cite{liu2023vectormapnet, Qiao_2023_CVPR} has also attracted considerable attention. NVAutoNet is well-equipped to extend its capabilities to address these tasks with ease.

\textbf{Multi-camera Fusion and BEV Perception.} Monocular camera perception tasks are well-established in computer vision, but multi-camera perception has recently gained attention \cite{Saha2021,Saha2022, Wang2021, Can2021StructuredBT,Reiher2020,Philion2020,Ng2020BEVSegBE}. Current methods, including ours, have shifted toward mid-level fusion, where information from various cameras is fused at the feature level. This demands a shared representation, often achieved through the BEV approach. BEV representation is advantageous for fusing data from multiple sensors and timestamps and can be directly used by downstream tasks like prediction, planning, and control.

\textbf{Perspective to 3D/BEV View Transformation.} Converting a perspective (image) view into a 3D/BEV view presents a challenging problem. Existing methods fall into four categories: homography-based, depth-based, MLP-based, and attention-based approaches. Homography-based methods (e.g., \cite{Reiher2020}) assume a flat-world model to shift pixels from a perspective view to a BEV view. However, this assumption often doesn't hold in real-world autonomous driving scenarios. Depth-based methods (e.g., \cite{Schulter2018, Philion2020, Reading_2021_CVPR}) require per-pixel depth information for transformation, but they face challenges. Inaccurate predicted depth can lead to poor-quality 3D feature maps, and the method's efficiency is hindered by high-dimensional 3D voxel feature maps. MLP-based methods (e.g., \cite{Chitta2021NEATNA, Yang2021, Mani2020MonoLO, Pan2020CrossViewSS, Hendy2020FISHINGNF}) are popular due to MLP's ability to transform data between spaces. However, these methods are camera-dependent and neglect key geometric information like camera intrinsic and extrinsic parameters, limiting their application to new sensor configurations. Their computational expense stems from stretching 2D feature maps into 1D vectors and performing a full connection operation. Recent work (e.g., \cite{Roddick2020PredictingSM, Saha2021, Saha2022}) alleviates computational demands by independently applying MLP operations to image columns. In contrast, attention-based methods (e.g., \cite{Saha2022, Can2021StructuredBT, Wang2021}) work backward (3D to 2D). They construct 3D or BEV queries that cross-attend to image features for 3D or BEV feature creation. While effective, these methods can be costly for cost-sensitive autonomous driving systems, especially with dense queries. For a more comprehensive understanding of these view transformation methods, refer to \cite{ma2023visioncentric}.

In our quest to create a real-time self-driving car perception system, we use a MLP-based method for our 2D-to-BEV view transformation. However, we take a unique approach by independently elevating individual image columns, inspired by techniques found in \cite{Roddick2020PredictingSM, Saha2021, Saha2022}. Our method differs these references in several ways. Firstly, we don't assume that each image column corresponds to a BEV ray, which doesn't hold for wide-field-of-view cameras like fish-eye cameras. Secondly, we expand the application of these techniques to fuse BEV features from multiple cameras using camera intrinsic and extrinsic parameters. Lastly, to ensure efficiency, we avoid using attention-based layers (e.g., as in done \cite{Saha2022}), opting instead for smaller MLP layers. Furthermore, in contrast to methods such as LSS \cite{Philion2020}, which initially construct a 3D volumetric feature map from 2D feature maps and depth information and subsequently reduce it to a BEV feature map, our approach generates a BEV feature map directly from 2D features. As a result, it is more computational and memory efficient. This strategy reflects our goal of achieving a balance between accuracy and computational efficiency.

\section{NVAutoNet}
\subsection{Overview}
NVAutoNet takes a set of $N_{view}$ camera images  $\{I_i\}_{i=1}^{N_{view}}$, covering a full 360-degree view of the surroundings, along with camera parameters. These images undergo 2D encoders to derive image features, which are then transformed into BEV features. BEV features specific to each camera are combined into a unified BEV feature map, fed into a shared BEV encoder to extract advanced features. Subsequently, this encoded BEV map enters task-specific encoders to generate 3D outputs such as obstacle cuboids, parking spaces, and driveable spaces. Refer to Figure \ref{fig:nvautonet} for an overview of the NVAutoNet process.

\subsection{2D Image Feature Extractors}
Each \emph{unrectified} input image $\mathbf{I}$ with dimensions $W \times H \times 3$ undergoes feature extraction, producing multiple feature maps ${F^k}$ at various scales. These $F^k$ have dimensions $C \times \frac{H}{2^{k+1}} \times \frac{W}{2^{k+1}}$. Our image feature extraction utilizes customized convolutional architectures, meticulously tailored for real-time operation. Comprising CNN blocks, the CNN backbone employs specific parameters like kernel size, stride, channel count, and repetitions. Network configurations for distinct camera groups are detailed in Table \ref{table:cam_encoders}. These parameters are searched using hardware-informed neural architecture search \cite{wang2022gpunet}, ensuring an optimal balance between accuracy and speed, with the omission of residual connections for faster processing \cite{NEURIPS2020_657b96f0}. The coarser maps merge with finer ones through upsampling, yielding multi-level feature maps enriched with semantic content. In the following sections, we delve into our approach for converting these 2D image feature maps into BEV representations.
\begin{table*}[t]
\small
	\begin{tabular}{llllllll}
	\hline
	& Input size & Blocks & Kernel sizes & Strides & Repeats & Channels \\
	\hline
        Front cam encoder & 480x960 & 5  &  7-3-3-3-3 & 4-1-2-2-2 & 1-0-2-5-3 & 32-32-128-256-512\\
        Side cam encoder  & 480x960 & 5  &  7-3-3-3-3 & 4-1-2-2-2 & 1-0-2-3-3 & 32-32-128-192-512\\
        Fisheye cam encoder & 480x960 & 5  &  7-3-3-3-3 & 4-1-2-2-2 & 1-0-2-3-3 & 32-32-64-96-512\\
        BEV encoder & 64x360 & 3  &  3-3-3 & 1-2-2 & 4-4-4 & 64-128-256\\
	\hline
	\end{tabular}
	\centering
	\caption{Configurations of lightweight camera and BEV CNN backbones.}
	\label{table:cam_encoders}
\end{table*}

\subsection{Image-to-BEV Transformation and Fusion}
\subsubsection{BEV Plane and BEV Grid}
The BEV representation is a popular choice in self-driving for efficiently depicting object positions, sizes, and aiding behavior prediction and planning. This representation involves a plane aligned with the vehicle's central axis, perpendicular to the Z axis. We divide this BEV plane into discrete sections using a grid $\mathbf{G}^{bev}$ with dimensions $M \times N$.

\subsubsection{Polar vs Cartesian Coordinates} 
The resolution of the BEV grid $\mathbf{G}^{bev}$ significantly impacts detection range and accuracy. For safe and comfortable autonomous highway driving, a detection range of 200 meters is ideal. However, using a regular Cartesian grid with cells of 0.25 meters results in a large $1600 \times 1600$ grid. This consumes excessive memory and computational resources, making it impractical to train deep neural networks and deploy them on vehicle chips like NVIDIA DRIVE Orin. To balance close and far-range resolution, we adopt an irregular BEV grid using Polar coordinates as similar to PolarNet \cite{Zhang_2020_CVPR}. This grid is defined by angular samples (ranging from 0 to 360) denoted by $M$, and depth samples represented by $N$. For instance, by employing 1-degree angular resolution and logarithmic radial spacing, we efficiently represent the BEV plane with a compact $360 \times 64$ grid.

\subsubsection{2D to BEV View Transformation}
\begin{figure*}
\centering
\includegraphics[width=0.45\textwidth]{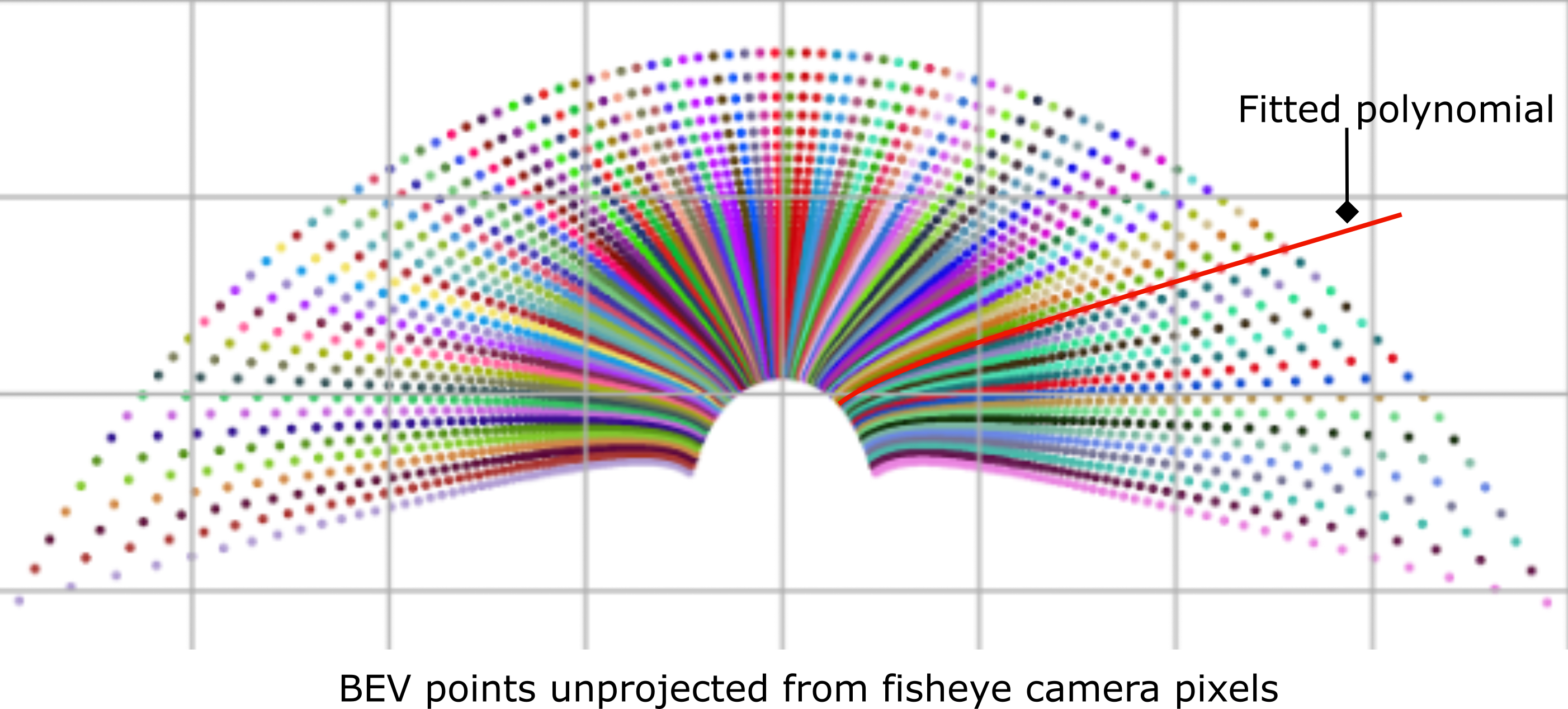} \hspace{1cm}
\includegraphics[width=0.45\textwidth]{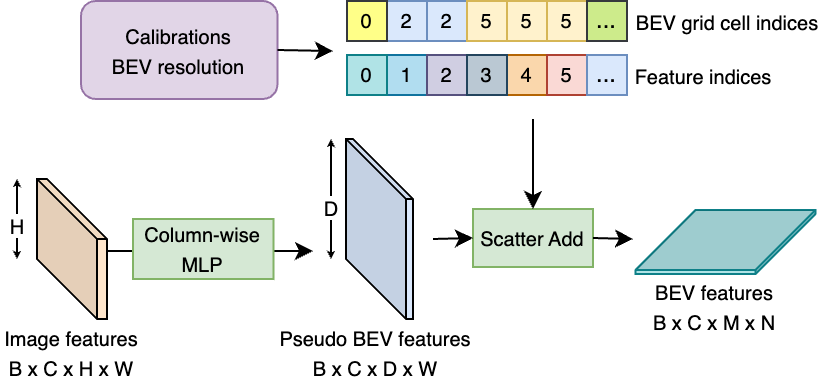}
\caption{\label{fig:bev_transformer} An overview of perspective to BEV view transformation. Left: Camera pixels are projected onto the BEV plane using camera intrinsic and extrinsic parameters. The resulting polar BEV points are then used to fit polynomial functions (one for each image column). These polynomial functions accept BEV radial distances as inputs and output corresponding BEV angular positions. Right: Image features are transformed into pseudo BEV features which are then transformed to the BEV features using BEV indices. The BEV indices are pre-computed using the fitted polynomial functions and a pre-defined BEV grid.} 
\end{figure*}

From a collection of 2D image feature maps originating from various cameras, our objective is to derive a singular BEV feature map, denoted as $F_{bev}$. In our approach, we adopt a data-centric strategy wherein we train a camera-to-BEV transformation function using a multilayer perceptron (MLP) network. Unlike prior MLP-based view transformation methods, our approach notably incorporates camera intrinsic and extrinsic parameters both during training and inference phases. Consequently, once trained, the model is capable of robustly adapting to different sensor configurations. A visual overview of our image-to-BEV view transformation module is provided in Figure \ref{fig:bev_transformer}.

A common observation is that there is a strong geometric relationship between row and column positions in the image plane and radial and angular positions in the BEV plane. In particular, each image column will correspond to a \textbf{curve} passing through the camera center on the BEV plane (see Figure \ref{fig:bev_transformer} left). If camera images are rectified, these curves become polar BEV rays as utilized in the previous works \cite{Saha2022, Saha2021}. In this work, input images are not rectified because rectifying fisheyes camera images is often challenging due to nonlinear, non-uniform distortion. Our goal is to learn functions (represented by neural networks) to transform every image column features onto the BEV plane.

Formally, let $\mathbf{I}^c$ be a column from an image $\mathbf{I}$ (for brevity camera index is omitted), we first project every pixel location $\mathbf{p}_i=[u_i, v_i] \in \mathbf{I}^c$ onto the BEV plane (using camera intrinsic and extrinsic parameters), and convert them to polar coordinates (angles and distances), resulting in a list of polar BEV points $\{\mathbf{b}_i = [a_i, d_i]\}$. Here it is worth noting that we are not merely projecting pixels to the BEV plane for view transformation as we do not assume the world is flat. Instead, these polar BEV points are used to fit a polynomial curve (one per image column) $a=f^c(d)$ that allow us to deduce a BEV angular position from a BEV radial position. This is illustrated in Figure \ref{fig:bev_transformer} (left), where examples of projected BEV points and fitted curves are presented. These fitted curves implicitly encapsulate camera intrinsic and extrinsic information.

For a given camera with a maximum detection range of $r$ meters, we discretize the range $[0, r]$ into $D$ logarithmically spaced bins. Let $\mathbf{c} \in \mathbb{R}^{H \times C}$ represent the encoded features corresponding to the image column $\mathbf{I}^c$ (again we dropped the feature level for notation simplicity), we define a BEV transformation network $\mathcal{B}(.)$ to yield $\mathbf{b} = \mathcal{B}(\mathbf{c})$ where $\mathbf{b} \in \mathbb{R}^{D \times C}$. Yet, these features aren't true BEV features at this stage; they are pseudo BEV features that must be aligned and combined into the global BEV feature map $F_{bev}$.

Each $\mathbf{b}$ feature comprises $D$ vectors, each associated with a radial distance bin. Consequently, the BEV plane location of each cell in $\mathbf{b}$ is determined using its distance (center of the distance bin) and the fitted polynomial function $f^c$. The same procedure is extended to all image columns and images, facilitating the transformation of image features from the image plane to the BEV plane. To optimize computational efficiency, only one feature map level at stride 8 is transformed to the BEV plane per image.

\textbf{BEV Pooling.} Leveraging a predefined BEV grid $\mathbf{G}^{bev}$, we generate the final BEV feature map $F_{bev}$ by combining all the computed BEV feature points described earlier. Specifically, the BEV feature points that belong to the same BEV grid cell are directly added together.

\textbf{BEV Indexing.} Assuming accurate camera calibration parameters, we can treat them as constants. In this context, we have the option to pre-calculate BEV indices, facilitating the mapping of pseudo BEV features to corresponding BEV grid cells. This BEV look-up-table expedites the image-to-BEV transformation process, ensuring remarkable efficiency for both training and inference operations. The computational overhead is notably minimal, requiring just 0.3ms per image.

\subsubsection{BEV Transformation using MLP}
The BEV transformation function $\mathcal{B}(.)$ is modeled using an MLP block with only one hidden layer, and the MLP parameters are shared by different image columns. Unlike convolutional layers, MLP layers have a capability to encode global contextual information (aka global attention), which we found very crucial for assigning image features into correct BEV positions as depth information has been lost and objects appears at different heights. 
\subsection{BEV Feature Extractor}
Similar to 2D image feature extractors, we adopt a CNN backbone to extract high-level features from the fused BEV feature map. See Table \ref{table:cam_encoders} for network configration details. The output of the BEV feature extractor, dubbed $\hat{F}_{bev}$, will be consumed by different 3D detection heads.
\section{Perception Tasks}
\subsection{3D Object Detection}
3D object detection is a key capability for autonomous driving. The goal is to localize, classify and estimate dimensions and orientations of objects in 3D space. Each object is represented by its category and 3D cuboid. In particular, each 3D cuboid has 9 degree-of-freedom (DOF) representing position, dimension, and orientation. In this work, we adopt a set prediction approach \cite{Rezatofighi2022} to remove the need for a non-maximum suppression (NMS) post-processing. 

The 3D object detection network includes five lightweight heads (implemented by a couple of convolutional layers), which takes the bottleneck feature map $\hat{F}_{bev}$ as input and predicts object class distributions and 3D cuboid parameters. Formally, let us denote $C \times M \times N$ be the dimension of $\hat{F}_{bev}$, where $M \times N$ is the spatial dimension and $C$ is the number of channels, the 3D detection network will output $\hat{K} = M \times N$ objects --- one object per grid cell. The model employs one head to predict the object classification scores, and three other heads for 3D cuboid parameters (position, dimensions and orientation) regression. There is an additional head for predicting uncertainty of cuboid parameters.\\
\textbf{Classification.} For $k$ number of objects, the network outputs $k+1$ classification channels, where the first channel represents object existence and the other $k$ channels represent a categorical distribution over $k$ classes. \\
\textbf{Position.} The network predicts a tuple $[r, a, e]$, where $r$ is radial distance, $a$ is azimuth angle, and $e$ is elevation. Note that the network actually predicts radial and angular offset values, which are then added to grid cell positions to form final radial and angular positions.\\
\textbf{Dimensions.} The network predicts three scalars $[d_x, d_y, d_z]$ which are absolute values in meters. \\
\textbf{Orientation.} We represent object orientation using a full rotation matrix $\mathbf{R} \in \mathbb{R}^{3 \times 3}$ , as opposed to previous works only estimating yaw angle. However, rotation matrix prediction is nontrivial because not every $3 \times 3$ matrix is a valid rotation matrix. Here we propose to train the network to predict sine and cosine values of yaw ($\psi$), pitch ($\theta$), and roll ($\phi$) angles respectively, which are later used to construct a rotation matrix by applying matrix multiplication of three axis rotation matrices together.
\begin{equation}
R = R_{z}(\psi)R_{y}(\theta)R_{x}(\phi),
\end{equation}
\begin{equation}
\begin{aligned}
R_{z} = \begin{bmatrix}
\cos \psi & - \sin \psi & 0\\
\sin \psi & \cos \psi & 0\\
0 & 0 & 1
\end{bmatrix},
R_{y}(\theta) = \begin{bmatrix}
\cos \theta & 0 & \sin \theta\\
0 & 1 & 0\\
-\sin \theta & 0 & \cos \theta
\end{bmatrix},
R_{x}(\phi)=\begin{bmatrix}
1 & 0 & 0\\
0 & \cos \phi & - \sin \phi\\
0 & \sin \phi & \cos \phi
\end{bmatrix}.
\end{aligned}
\end{equation}

\textbf{Training Losses.} For each scene, let $\mathcal{G} = \{\mathbf{g}_i\}_{i=1}^K$
be  a set of $K$ ground-truth objects, and let $\mathcal{P} =\{\mathbf{d}_i\}_{i=1}^{\hat{K}}$ be a set of $\hat{K}$ predicted objects. The training loss is computed in a two-step fashion where in the first step, we found the best one-to-one matching between $\mathcal{G}$ and $\mathcal{P}$. In the second step, we compute the final loss based on the matching result. Although the idea sounds very simple, the training process will be highly affected by the matching quality --- if the matching algorithm returns wrong or sub-optimal matches, the network training won't be successful. 

To ensure high quality matches, we constructed a matching cost function considering classification, position, dimension, and orientation costs between each candidate-ground truth pair. In the early phase of training, however the network prediction is quite noisy, often leading to bad assignments. This issue can be mitigated by setting a higher weight for the position cost. Alternatively, inspired by \cite{peize2020onenet}, we proposed to limit the matching candidates for each ground truth object by its corresponding coverage on the BEV grid. This will prohibit matching pairs in which the ground truth object and the candidate are far apart. Moreover, the matching optimization problem is now simplified,  we adopt a greedy matching algorithm instead of the well-known Hungarian algorithm without losing accuracy, but more efficient.

Let $\mathcal{P}_{pos}$ be the set of positive object candidates matched to the ground truth objects, and $\mathcal{P}_{neg} = \mathcal{P} \setminus \mathcal{P}_{pos}$ be a set of negative objects, the training loss is as below:
\begin{equation}
\begin{aligned}
L_{obs}(\mathcal{G}, \mathcal{P}) = L^{pos}_{obs}(\mathcal{G}, \mathcal{P}_{pos}) + L^{neg}_{obs}(\mathcal{P}_{neg}) \\ = \sum_{\mathbf{d}_i \in \mathcal{P}_{pos}} L(\mathbf{g}_i, \mathbf{d}_i) + \sum_{\mathbf{d}_i \in \mathcal{P}_{neg}}L(\mathbf{d}_i),
\end{aligned}
\end{equation}
where $L(\mathbf{g}_i, \mathbf{d}_i)$, and  $L(\mathbf{d}_i)$ are loss functions per each predicted candidate. % $w_{\mathbf{g}_i}$ is a weighting parameter pre-computed for each ground-truth object $\mathbf{g}_i$. In this work, $w_{\mathbf{g}_i}$ is simply equal to the number of views in which the object $\mathbf{g}_i$ appear. Effectively, objects appearing in multiple views are expected to have higher detection accuracy.
If $\mathbf{d}_i$ is a negative candidate,  the loss function $L(\mathbf{d}_i)$ is simply a (focal) binary cross entropy loss --- pulling its objectness score to zero. If $\mathbf{d}_i$ is a positive candidate, the loss function $L(\mathbf{g}_i, \mathbf{d}_i)$ composes of classification and regression losses. While the classification loss is still a (focal) cross-entropy loss, the regression loss is more complex as explained below.

%Note that each ground-truth and prediction 3D cuboid has 15 parameters (3 for position, 3 for dimensions, and 9 for orientation), 
To regress 3D cuboid parameters, ideally we should compute a global loss such as intersection-over-union (IoU) score which considers all parameters together as there is a strong correlation between cuboid parameters. However there is no closed-form solution to compute IoU between two 3D cuboids. Here, we propose to decompose the 3D cuboid regression loss into position (location) loss, shape (size) loss and orientation (rotation) loss, as below:
\begin{equation}\label{eq:obs_reg}
L^{reg}(\mathbf{g}_i, \mathbf{d}_i) = L^{loc}(\mathbf{g}_i, \mathbf{d}_i) + L^{size}(\mathbf{g}_i, \mathbf{d}_i)  + L^{rot}(\mathbf{g}_i, \mathbf{d}_i) 
\end{equation}
\begin{equation}
\begin{aligned}
L^{loc}(\mathbf{g}_i, \mathbf{d}_i) =\frac{| \mathbf{g}_i^r -  \mathbf{d}_i^r|}{\sigma_r} + \frac{| \mathbf{g}_i^a -  \mathbf{d}_i^a|}{\sigma_a} + \frac{| \mathbf{g}_i^e -  \mathbf{d}_i^e|}{\sigma_e} \\+ \log(2\sigma_r) + \log(2\sigma_a) + \log(2\sigma_e)
\end{aligned}
\end{equation}
\begin{equation}
L^{size}(\mathbf{g}_i, \mathbf{d}_i) =\frac{1}{\sigma_s} (1 - \prod_{d \in \{d_x, d_y, d_z\}} \frac{\min (\mathbf{g}_i^{d}, \mathbf{d}_i^{d})}{\max (\mathbf{g}_i^{d}, \mathbf{d}_i^{d})}) + \log(2\sigma^s)
\end{equation}
\begin{equation}
L^{rot}(\mathbf{g}_i, \mathbf{d}_i) = \frac{1}{\sigma_o} \sum_{r \in R} | \mathbf{g}_i^{r} - \mathbf{d}_i^{r}| + \log(2\sigma^o)
\end{equation} 
where $\sigma_r$,$\sigma_a$, $\sigma_e$, $\sigma_s$, $\sigma_o$ are  uncertainty values for position, shape and orientation respectively. These values are predicted by the network.

\subsection{3D Freespace} 
3D obstacle detection generally covers category-wise classifiable vehicles and vulnerable road users (VRU). In a driving scenario, there is a lot more information which is relevant for safe driving beyond the predictions of 3D obstacle detection. For example, there can be random hazard obstacles like tyres, traffic cones lying on the road. Additionally, there are a lot of static obstacles like road-divider, road-side curb, and guard rails which are not covered by 3D obstacle detection. An autonomous vehicle has to drive safely within the boundaries of the road by avoiding all kinds of obstacles. The region within the boundaries of the road which is not occupied by any obstacle could be carved out as a driveable region. The driveable region is used interchangeably as the freespace region. The down-stream behaviour planner would consume the freespace region information to plan a safe trajectory for the AV. So it is essential to have a perception component which predicts this freespace region. The freespace region is represented as a radial distance map (RDM). The representation is explained in more details in the next section.

\textbf{Radial Distance Map.} While polygons are used for labeling of 3D freespace, we use the RDM representation due to its higher efficiency. RDM is composed of equiangular bins and radial distance values for each angular bin to denote spatial locations. For autonomous driving applications, the distance to the closest freespace boundary is the most important one and thus we use a single scalar for each angular bin to represent the closest freespace boundary. In order to create ground-truth RDM for 3D freespace, we simply shoot a ray from the center of BEV plane at each angular bin direction and compute the intersection between the ray and the ground-truth polygon. The most important benefit of using RDM to represent 3D freespace is it can be directly used to create 3D boundary points without additional post-processing. 
In addition to the radial distance, we also have, for each angular bin, boundary semantic labels such as vehicle, VRU and others. 
Each freespace ground truth label becomes $(\mathbf{r}, \mathbf{c})$, where $\mathbf{r}$ is a radial distance vector, and $\mathbf{c}$ is a boundary semantic vector.

\textbf{Model.} The 3D freespace detection network consists of a shared neck and two separate heads. The shared neck includes a couple of convolutional layers on top of the bottleneck feature map $\hat{F}_{bev}$. It is later extended into two heads which predict radial distance and classification maps.

\textbf{Training Losses.} For each scene, let $\mathcal{G} = (\mathbf{r}, \mathbf{c})$ be the ground-truth, let $\mathcal{P} = (\hat{\mathbf{r}}, \hat{\mathbf{c}})$ be the prediction.
%Here, ˆd={^di}i=1,…,N\hat{d} = \{\hat{d_i}\}_{i=1,\ldots,N} is the prediction RDM from the radius head and d={di}i=1,…,Nd = \{d_i\}_{i=1,\ldots,N} is the label RDM. Assuming we do KK-class classification in each of the angular bins, classification label is defined as c={ci}i=1,…,Nc = \{c_i\}_{i=1,\ldots,N} where ci∈[0,K)c_i \in [0, K) is the corresponding label class index. If pijp_{ij} is the probability from the classification head corresponding to ithi^{th} angular bin and the jthj^{th} class, the classification prediction is defined as ˆc={{pij}j=1,…,K}i=1,…,N\hat{c} = \{\{{p_{ij}\}_{j=1,\ldots,K}}\}_{i=1,\ldots,N}.
 The overall loss function for the 3D freespace detection task is defined as:
\begin{equation}
L_{fsp}(\mathcal{G}, \mathcal{P}) = L^{reg}_{fsp}(\hat{\mathbf{r}}, \mathbf{r}) +  L^{cls}_{fsp}(\hat{\mathbf{c}}, \mathbf{c})
\end{equation}
\begin{equation}
L^{reg}_{fsp}(\hat{\mathbf{r}}, \mathbf{r}) = L^{iou}_{fsp}(\hat{\mathbf{r}}, \mathbf{r}) + L^{sim}_{fsp}(\hat{\mathbf{r}}, \mathbf{r})
\end{equation}
where $L^{reg}_{fsp}(\hat{\mathbf{r}}, \mathbf{r})$ is the regression loss which is a combination of radius loss $L^{iou}_{fsp}(\hat{\mathbf{r}}, \mathbf{r})$ and the similarity loss $L^{sim}(\hat{\mathbf{r}}, \mathbf{r})$; $L^{cls}_{fsp}(\hat{\mathbf{c}}, \mathbf{c})$ is the classification loss.
The radius loss is a polar intersection-over-union loss (IoU) computed between the prediction and ground truth labels, defined as below:
\begin{equation}
L^{iou}_{fsp}(\hat{\mathbf{r}}, \mathbf{r})  = 1.0 - \prod_{i=1}^{N_{bins}} \frac{\min(r_i, \hat{r_i})}{\max(r_i, \hat{r_i})}.
\end{equation}
The similarity loss is computed between the line segments formed by joining the end points from the consecutive angular bins in the prediction and ground truth labels. This loss helps in reducing the noise in the predicted RDM.
\begin{equation}
L^{sim}_{fsp}(\hat{\mathbf{r}}, \mathbf{r}) = \sum_{i=1}^{N_{bins}} (1.0 - \frac{{\hat{l}}_{i}^{i+1} \cdot {l}_{i}^{i+1}}{\left\lVert {\hat{l}}_{i}^{i+1} \right\rVert \left\lVert {l}_{i}^{i+1} \right\rVert}),
\end{equation}
where ${\hat{l}}_{i}^{i+1}$ is the line segment formed by joining the end points of $\hat{r}_{i}$ and $\hat{r}_{i+1}$. ${{l}}_{i}^{i+1}$ is defined similarly.
Finally, the classification loss $L^{cls}_{fsp}(\hat{\mathbf{c}}, \mathbf{c})$ is the standard focal loss, i.e., 
\begin{equation}
L^{clc}_{fsp}(\hat{\mathbf{c}}, \mathbf{c}) = \sum_{i=1}^{N_{bins}} \sum_{j=1}^{C} (1 - p_{ij})^{\gamma} \log(p_{ij}),
\end{equation}
where $\gamma$ is the focal loss parameter.
\subsection{3D Parking Space}
Another important aspect of autonomous driving is the ability to localize and classify parking spaces. Each parking space is represented as an oriented rectangle, parameterized by $[cx, cy, l, w, \theta]$, where $cx$ and $cy$ are the center coordinates of the box, $l$ and $w$ are the length and the width of the box in meters respectively, and $\theta$ is the orientation of the box (yaw angle in radians) in the range $[0, \pi)$. Note that an oriented box angled at $\pi$ visually appears the same as a box oriented at 0, thus the orientation value $\theta$ need not cover the entire angular range of $[0, 2\pi)$.
Knowing the \textit{profile} of every parking space is important for planning and control purposes. As such, every prediction output by our model will be assigned a parking profile. In the current system we support three different parking profiles: \textit{angled, parallel and perpendicular}. As their name suggests, the profiles denote the types of planning and control maneuvering required to successfully park the car in the parking space. \textit{Parallel} parking spaces are ones that typically appear on the side of the street and require a parallel parking maneuver. Conversely, \textit{angled} and \textit{perpendicular} parking spots are ones where the car can be parked straight in (or backed in). 

\textbf{Model.} The parking space detection task follows the same design (head, training strategy and losses) of the obstacle detection task. The parking detection network consists of classification and regression heads. The classification head predicts per-profile confidence scores. We rely on each parking spot's profile to implicitly encode its existence score. The regression head predicts the parking space oriented bounding boxes as discussed above.

\textbf{Training Losses.} The training loss is similar to that of the obstacle detection task, but its regression loss function is much simpler. Let $\mathbf{g}_i$ and $\mathbf{d}_i$ be a matched pair of ground truth and detection, the regression loss is defined as:
\begin{align}
    \loss^{reg}_{prk}(\mathbf{g}_i, \mathbf{d}_i) = \sum_{s \in \{cx, cy, l, w, \theta\}} ({\mathbf{g}_i^{s} - \mathbf{d}_i^{s}})^2.
\end{align}

\section{Multi-task Learning and Loss Balancing}\label{sec:loss_balancing}
NVAutoNet functions as a multitask network, adeptly handling various tasks simultaneously. These tasks inherently vary, making a uniform training approach less optimal. Additionally, distinct task losses occupy different terrains. A simple combination of loss components with equal weights might underemphasize certain tasks or let some tasks dominate the overall loss. 

\textbf{Adaptive Weight Adjustment}: With $T$ number of tasks, the combined training loss for batch $b$ is $L_{total} = \sum_{t=1}^{T} w_t L_{t}^b$, where $w_t$ is the task-specific loss weight, and $L_t^b$ is the task loss for batch $b$. We introduce a straightforward yet powerful algorithm to dynamically modify the loss weights $\{w_t\}_{t=1}^T$ for each loss component. In a dataset of $S$ samples, $L_{t,s}$ represents the loss for task $t$ on sample $s$. At the start of each epoch, the total losses for each task $t$ across all samples are computed. The loss weight $w_t$ is then set as the reciprocal of this loss sum, proportionally adjusted by a predefined task loss prior $c_t$, i.e.,
\begin{equation}
L_{t} = \sum_{s=1}^{S} L_{t,s}, \quad \hat{w}_t = \frac{c_t}{L_t}, \quad w_t = \frac{\hat{w}_t}{\sum_{t=1}^{T} \hat{w}_t}.
\end{equation}
$c_t$ is a configurable loss multiplier for task $t$ that can be used to boost or reduce a task's loss. The approach of scaling losses by their reciprocal loss sums intuitively aligns all losses on a comparable scale, thereby minimizing disparities in gradient magnitudes. The inclusion of loss multipliers ${c_t}$ becomes valuable when specific tasks are either more significant or present higher complexity. 

In the initial epoch, we manually set all task loss weights $w_t = 1$ for all $t \in [1, T]$, since we lack previous epoch's loss sum data. After each epoch, we iteratively update $w_t$, ensuring a progressive refinement process.

\textbf{Two-Stage Approach}: Our approach unfolds in two stages. Initially, in the first training round, we uniformly establish ${c_t}$ values at 1. Subsequently, we assess the outcomes of multi-task training in comparison to single-task training. By gauging improvements or declines in key performance indicators (KPIs) for individual tasks, we tune suitable ${c_t}$ values. Equipped with these ${c_t}$ values, we proceed to retrain the network in subsequent iterations. Notably, while ${c_t}$ values are fine-tuned manually, our observations suggest that the pursuit of optimal ${c_t}$ values is notably simpler compared to directly searching for ${w_t}$ values.

\section{Experimental Evaluation}
This section highlights NVAutoNet's latency and accuracy achievements. Ideally, comparing NVAutoNet to state-of-the-art methods would be insightful. However, we encountered various challenges. Firstly, while there exist BEV object detection methods and benchmark datasets, freespace and parking space detection are rare. Secondly, many existing BEV perception methods, like 3D object detection, prioritize accuracy, unlike NVAutoNet which optimizes for both accuracy and latency. Thirdly, NVAutoNet is tailored for real self-driving applications with a required detection range of up to 200 meters. This makes some design choices (like image to BEV view transformation, polar BEV representation) less favorable for public datasets like nuScenes, which primarily consider shorter ranges (less than 70 meters). Consequently, our evaluation predominantly focuses on our in-house dataset and the NVIDIA DRIVE Orin SoC platform.
\subsection{Datasets and Evaluation Metrics}
\subsubsection{Datasets}\label{sec:datasets}
Our in-house datasets consist of real, simulated reality and augmented reality data. In total, there are 2.2M training scenes, 400K validation scenes and 177K testing scenes. Table \ref{table:datasets} summarizes our datasets. Lidar data was used to generate ground truth labels. Our data contains a fair amount of noisy labels due to view point differences between Lidar and camera sensors. For example, Lidar is mounted at a higher position than cameras, thus there are obstacles, which may be visible by Lidar, are hardly visible by cameras. These issue not only affect model training but also model evaluation (e.g., low recall rates).
\begin{table}
	\begin{tabular}{lcc}
	\hline
	& Amount \\
	\hline
	Number of cameras per scene & 8  \\
        \hline
        Real training samples & 2M \\
        Sim training samples & 200k \\
        Validation samples & 400K \\
        Test samples & 177K \\
        \hline
        Number of countries & 20  \\
        \hline
        Number of obstacle classes & 5 \\
        Number of freespace classes & 3 \\
        Number of parking classes & 3 \\
        \hline
        Percentage of dry roads & 95.17$\%$ \\
        Percentage of wet roads & 4.83$\%$  \\
        \hline
        Bright light condition & 49.5$\%$   \\
        Diffused light condition & 31.4$\%$   \\
        Poor light condition & 19.1$\%$   \\
	\hline
	\end{tabular}
	\centering
	\caption{In-house dataset summary.}
	\label{table:datasets}
\end{table}
\begin{table}[h!]
	\begin{tabular}{llcl}
	\hline
	Components & Latency (ms)  &Total (ms)\\
	\hline
	Front Cam Encoder  &  1.80 $\times$ 2   & 3.61\\
	Side Cam Encoder   &  1.60  $\times$ 2 & 3.21\\
    Fisheye Cam Encoder&  1.39  $\times$ 4 & 5.58\\
    3D Uplifting + Fusion   &  0.30      $\times$ 8 & 2.40 \\
    3D Encoder + Heads &  3.91  $\times$ 1 & 3.91\\
        &   &   18.72 \\
	\hline
	\end{tabular}
	\centering
	\caption{NVAutoNet latency (ms) measured on NVIDIA DRIVE Orin embedded GPU. The model runs at 53 FPS.}
	\label{table:latency}
\end{table}
\subsubsection{Evaluation Metrics}
\textbf{3D Obstacles.} 
We calculate obstacle detection metrics based on identifying true positives (TP), false positives (FP), and false negatives (FN) from detection outputs and ground truth labels. For each class, we find one-to-one matching between detection output and ground truth using greedy algorithm with the Euclidean distance between their centroids. A match is valid if the relative radial distance between a prediction and ground truth objects is less than 10$\%$, and their absolute azimuth error is less than 2 degrees.  
All unmatched detections become FP while all unmatched ground truth becomes FN. Once TP, FP, and FN have been identified, we compute precision, recall, F1-score, AP and mAP KPIs. Moreover, we search for the best confidence threshold that maximizes F1-score, and compute regression errors for all true positive detections. \emph{Position error} measures relative radius error ($\%$), absolute azimuth error (degrees) and absolute elevation error (meter). \emph{Orientation error} is defined as $|| \textrm{log}(R^{-1}\hat{R})||$, where $R$ and $\hat{R}$ are ground truth and prediction rotation matrices. \emph{Shape error} measures relative error for length, width, and height ($\%$). We also define the safety mAP based on a safety zone. The safety zone is defined as a rectangular region around the ego vehicle, i.e., 100 meters ahead and behind the ego vehicle and 10 meters left and right of the vehicle. 

\textbf{3D Freespaces.}
%As freespace regression is represented as radial distance maps, evaluating 3D freespace output accuracy simply computes the errors between ground truth and prediction radial distances for each angular bin. Similarly freespace classification related errors are computed between ground truth and prediction classifications.\\
Given a pair of ground truth and prediction freespace RDMs, we compute the following metrics (averaged over angular bins and frames). \emph{Relative gap} measures the relative radial distance errors ($\%$). 
\emph{Absolute gap} measures the absolute radial distance errors (meters). \emph{Success rate} measures the percentages of successfully estimated angular bins. Angular bins are considered as successfully estimated when the relative gap is less than $10\%$. \emph{Smoothness} measures the total variation of radial distance maps defined as $\sum_{i=1}^{N_{bins}} |r_i - r_{i-1}|$. \emph{Classification error} measures precision and recall for each label class.

\textbf{3D Parking Spaces.} Similar to object detection, we compute precision, recall, F1 and AP metrics.  Intersection over union (IoU) scores are used to match predictions to ground truth labels. A match is valid if the IoU $\ge$ 70\%. This strict criteria is necessary for real-world applications of autonomous parking as small misalignment between the detection and the actual parking space position can lead to imperfect parking. We also compute mean IoU values for all true positive detections.

\subsection{Latency Performance}
%We export NVAutoNet using NVIDIA TensorRT and time it on NVIDIA Orin. Table \ref{table:latency} reports the total NVAutoNet latency as well as the breakdown latency numbers for different NVAutoNet components. It can be seen that our network only costs \textbf{18.7 ms} (8 cameras input), leading to a very high frame-rate at 53 FPS. Compared to BEVDET \cite{huang2021bevdet}, one of the most efficient BEV object detection methods, NVAutoNet is 7.5x times faster even though BEVDET's latency was measured on NVIDIA GeForce RTX 3090 GPU, which is much more powerful than NVIDIA Orin SoC. Compared to Fast-BEV \cite{fastbev}, a method that is highly optimized for real-time applications, NVAutoNet is still a lot faster (53 vs 45 FPS). Both NVAutoNet and Fast-BEV are measured on the same NVIDIA Orin SoC. It is also worth mentioning that NVAutoNet handles multiple tasks jointly and its BEV perception range is much higher than those of the competitors (e.g., 200 vs 50 meters).
NVAutoNet is deployed using NVIDIA TensorRT and evaluated on the NVIDIA DRIVE Orin platform. The total latency and individual latency figures for distinct NVAutoNet components are presented in Table \ref{table:latency}. Notably, our network operates at an impressively low duration of just \textbf{18.7 ms} (for an 8-camera input), enabling a remarkable frame rate of 53 FPS.

In comparison to BEVDET \cite{huang2021bevdet}, a highly efficient BEV object detection method, NVAutoNet showcases a 7.5x speed boost, even though BEVDET's latency measurement was conducted on the more potent NVIDIA GeForce RTX 3090 GPU, surpassing the NVIDIA DRIVE Orin SoC's capabilities. Compared to Fast-BEV \cite{fastbev}, an approach highly tailored for real-time use cases, NVAutoNet still outperforms with 53 FPS against Fast-BEV's 45 FPS. Both NVAutoNet and Fast-BEV were evaluated on the identical NVIDIA DRIVE Orin SoC.

\subsection{Qualitative Performance}
\begin{figure*}
\centering
\begin{subfigure}[b]{.375\textwidth}
    \includegraphics[width=1.0\textwidth]{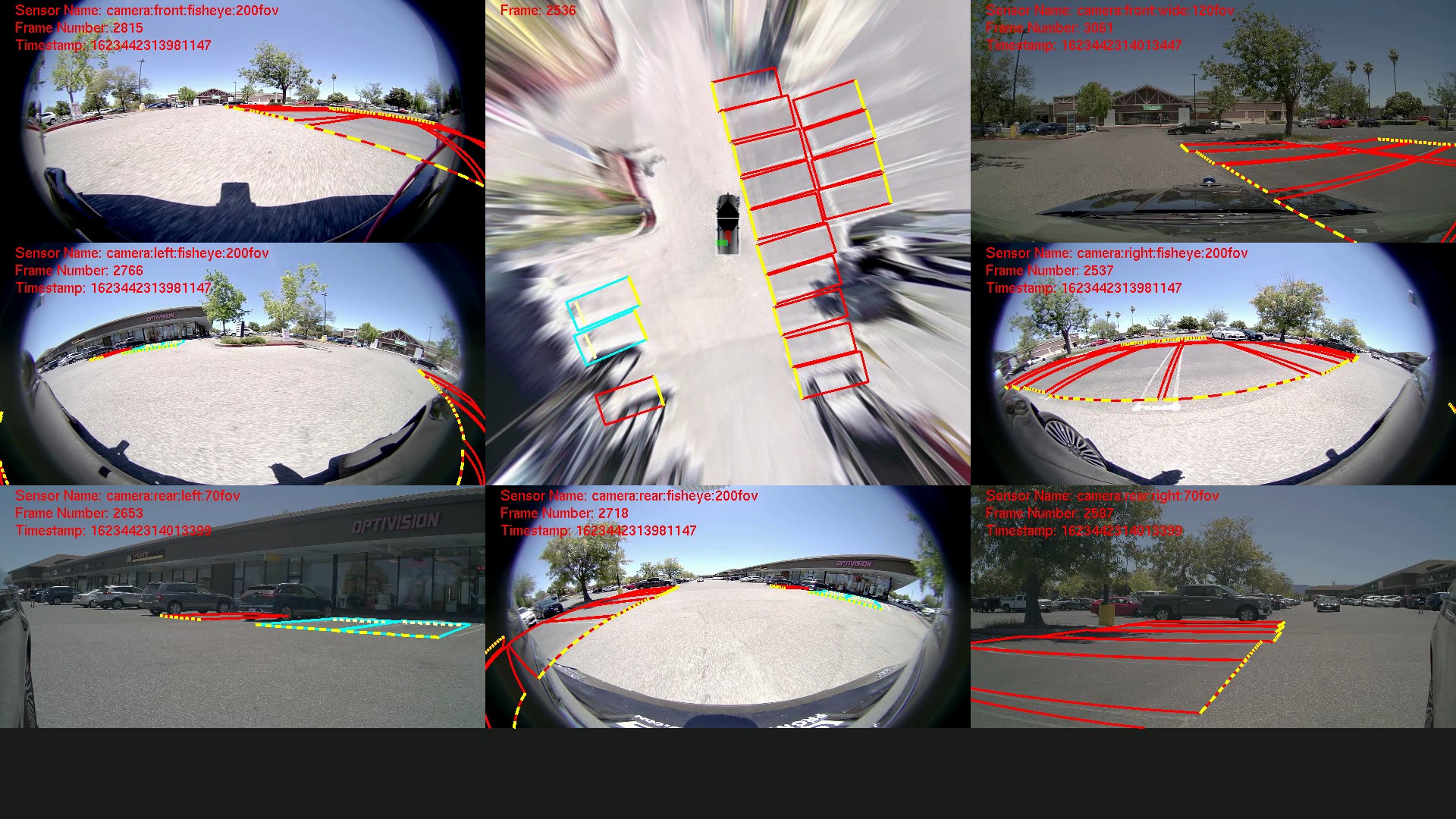}
\end{subfigure}
\begin{subfigure}[b]{.3\textwidth}
    \includegraphics[width=1.0\textwidth]{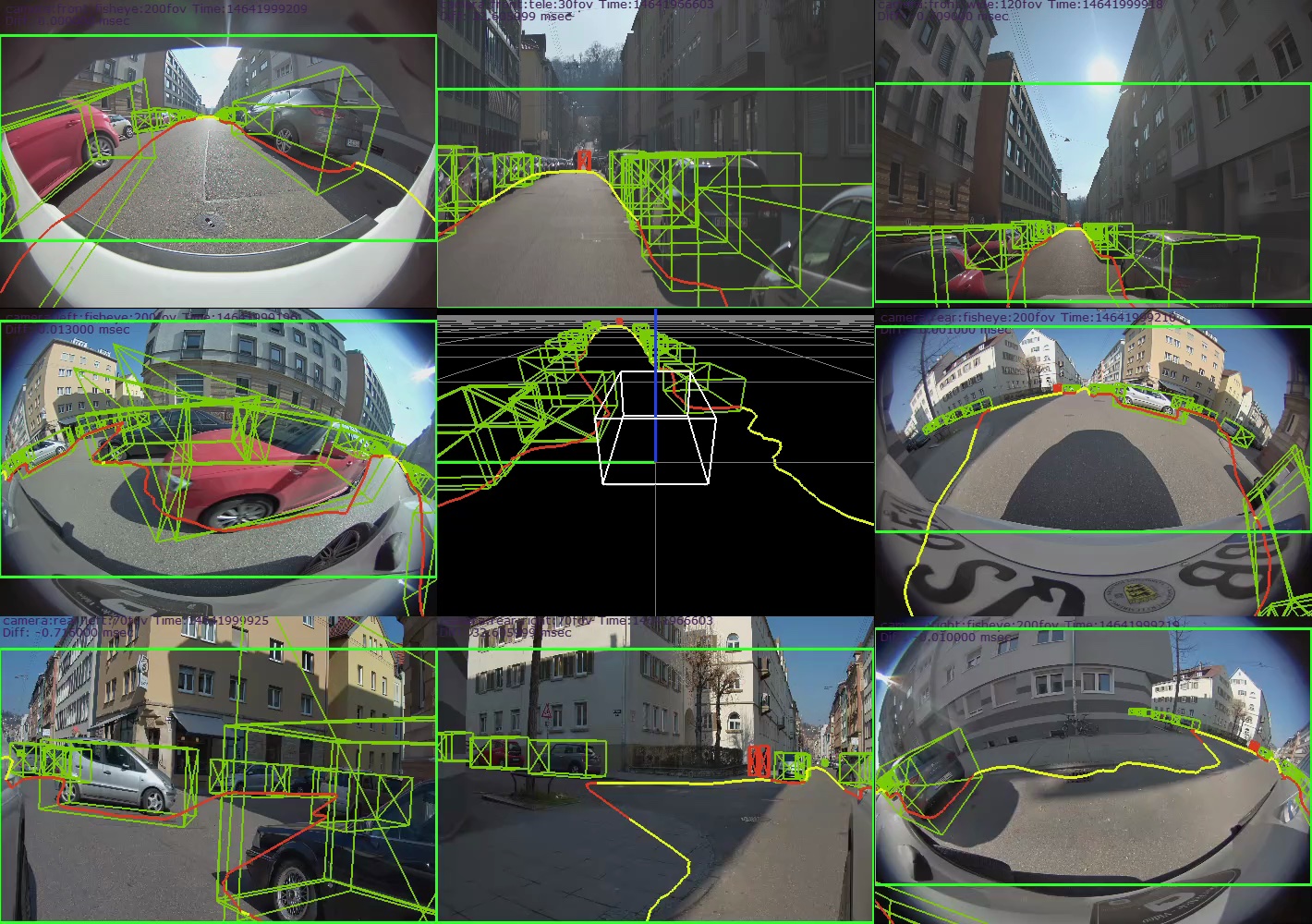}
\end{subfigure}
\begin{subfigure}[b]{.3\textwidth}
    \includegraphics[width=1.0\textwidth]{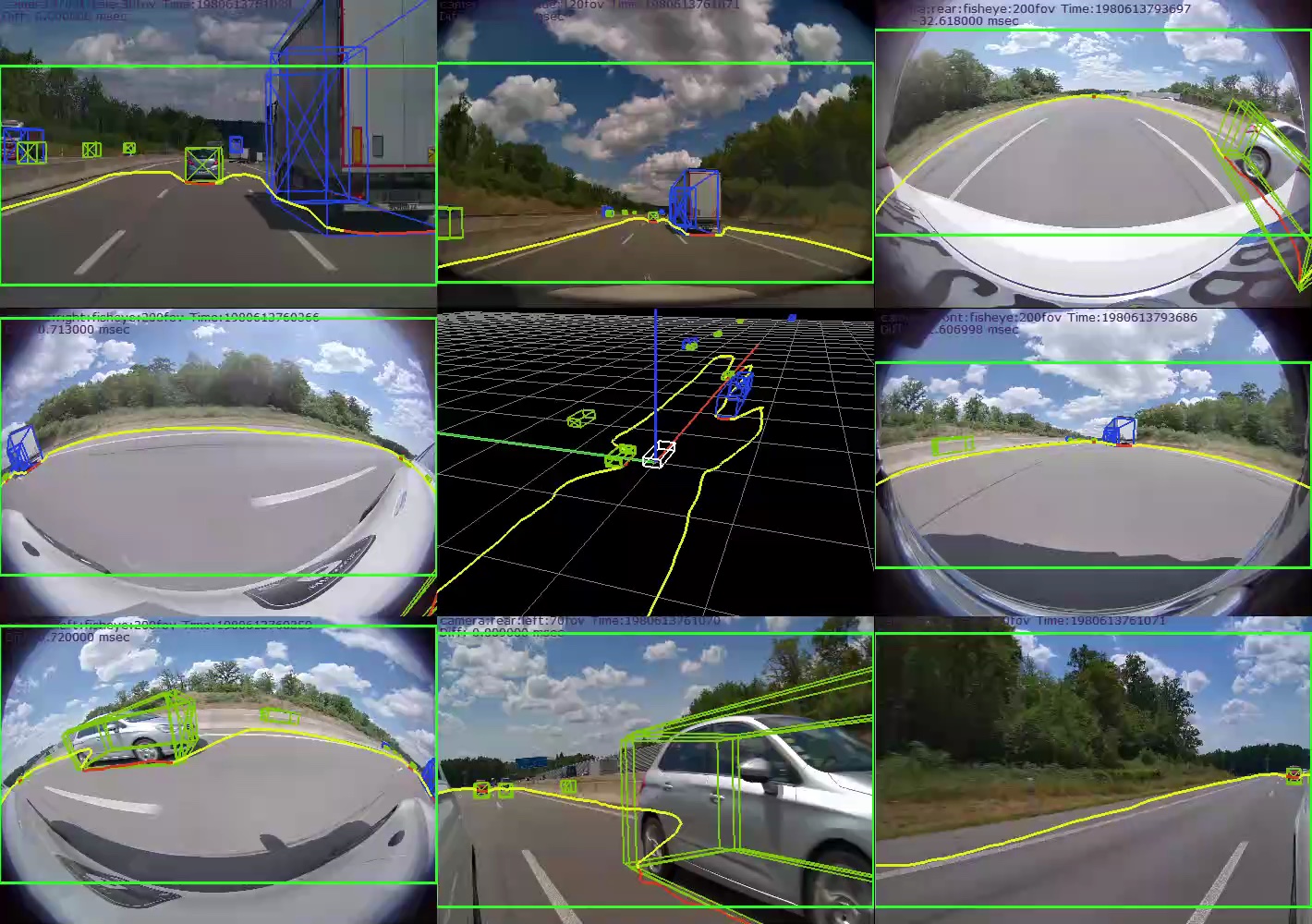}
\end{subfigure}
\caption{NVAutoNet's qualitative results. Left: Parking space detection in a parking lot. Middle: Obstacle and freespace detection in an urban area with a sedan car. Right: Obstacle and freespace detection in a highway area with a suv car.}
\label{fig:autonet_vis_results}  
\end{figure*}

Figure \ref{fig:autonet_vis_results} shows visual results of a single NVAutoNet model tested on different scenarios (i.e., parking lot, urban and highway) and different car models (i.e., Sedan and SUV). More qualitative results can be found in \href{https://youtu.be/cPxVhCJ7kyY}{the video}. 

\subsection{Quantitative Performance}
\subsubsection{3D Obstacles} \label{sec:results_3d_obs}
Presented in Table \ref{table:obstacle_map}, the obstacle detection accuracy outcomes are showcased. The overall mAP score achieves 0.465. Notably, the network excels in detecting vehicles, attaining an AP score of 0.648, while it faces challenges in identifying persons/pedestrians, with an AP score of 0.351. This observation aligns with the typical size discrepancy between persons and vehicles. When focusing solely on a safety region (within 100 meters ahead and behind the ego vehicle, and 10 meters to its sides), the mAP score notably improves to 0.595.

The network demonstrates reasonable accuracy in estimating orientations for vehicles and trucks, displaying average errors below 7 degrees. However, orientation errors for bikes and persons are relatively high, averaging at 12.3 and 53.4 degrees, respectively. For more detailed insights into detection results, Table \ref{table:obstacle_f1_range} provides a granular breakdown. Generally, detection accuracy drops steadily as distances increase.
\begin{table}[t]
	\begin{tabular}{lccccccc}
	\hline
	 & Vehicle & Truck & Person  & Bike-rider & mAP & (safety) mAP \\
	\hline
	AP & 0.638 & 0.388 & 0.351 & 0.483 & 0.465 & 0.595 \\
        Position error & 0.989 & 1.932 & 0.614 & 0.7 & - & - \\
        Orientation error & 6.542 & 5.295 & 53.4 & 12.334 & - & - \\
	\hline
	\end{tabular}
	\centering
	\caption{Obstacle detection accuracy for different classes.}
	\label{table:obstacle_map}
\end{table}
\begin{table}
	\centering
	\begin{tabular}{llcccc}
	\hline
	Class & Range (meters) & F1-score & Position errors & Orientation errors & Shape errors \\
	\hline
	Vehicle & 0-50    & 0.715 &	0.741 &	7.213 & 0.173 \\
        Vehicle & 50-100  & 0.481	& 2.231 &	11.295 & 0.246 \\
        Vehicle & 100-150    & 0.371 &	4.098 &	12.304 & 0.306 \\
        Vehicle & 150-200  & 0.250	& 6.647	 & 9.193  & 0.339\\
        \hline
        Truck & 0-50    & 0.481 &	1.099	&  8.463 & 0.362 \\
        Truck & 50-100  & 0.430 &	2.269	& 7.064 & 0.453\\
        Truck & 100-150    & 0.419	& 3.554	& 6.854 & 0.420 \\
        Truck & 150-200  & 0.386	& 5.334	 & 6.778 & 0.411\\
        \hline
        Pedestrian & 0-30   & 0.491 &	0.500 &	37.772 & 0.435  \\
        Pedestrian & 30-50  & 0.413	& 1.259 &	51.483 & 0.480 \\
        Pedestrian & 50-100 & 0.313	& 2.283	& 60.708 & 0.528 \\
        \hline
        Bike-with-rider  & 0-30  & 0.574	&	0.437	&	11.876 & 0.297 \\
        Bike-with-rider & 30-50  & 0.482	&	1.152	&	16.516 & 0.352 \\
        Bike-with-rider & 50-100 & 0.387	&  2.055 &	19.197 & 0.428  \\
	\hline
	\end{tabular}
	\caption{Obstacle detection accuracy at different radial distance ranges.}
	\label{table:obstacle_f1_range}
\end{table}
%\begin{table}[hbt!]
%	\begin{tabular}{llcccc}
%	\hline
%	Model & Datasets & Range (m) & mAP & FPS & Latency\\
%	\hline
%	NVAutoNet  &  In-house & 200  & 0.46 & 53.4 & 18.7\\
%        Fast-BEV-tiny   &  nuScenes eval & 50 & 0.277 & 44.8 & 22.3\\
%	Fast-BEV-small   &  nuScenes eval & 50 & 0.369 & 16.6 & 60.3\\
%	\hline
%	\end{tabular}
%	\centering
%	\caption{NVAutoNet versus Fast-BEV \cite{fastbev}. This is not a apple-to-apple comparison but a good reference.}
%	\label{tab:autonet_vs_fastbev}
%\end{table}
\subsubsection{3D Freespace}
Table \ref{table:freespace_result_solo} offers a comprehensive overview of the freespace evaluation outcomes. For more detailed insights, Table \ref{tab:freespace_result_sector} presents segmented regression metrics across various angular and radial sectors. Notably, regions in close proximity demonstrate higher accuracy compared to distant ones. Moreover, the front area exhibits greater accuracy than the rear section. This discrepancy arises from the fact that the front region benefits from coverage by three distinct cameras: 120FOV, 30FOV, and fisheye 200FOV.

Notably, precision and recall for the "Other" and "Vehicle" categories surpass those of the "VRU" category. This variance can be attributed to the inherent complexity of VRU classification. VRUs are often captured within fewer angular bins compared to other categories, contributing to the increased difficulty in their precise classification.

%Table \ref{table:freespace_result_solo} summarizes the overall freespace results. Additionally, Table \ref{tab:freespace_result_sector} reports bucketized regression metrics per sector divided into different combinations of angular and radial sectors. Overall, close range regions have higher accuracy than far range regions. Also front region is more accurate than rear region. This is because the front region is covered by three different cameras: 120FOV, 30FOV and fisheye 200FOV. Also it can seen that precision and recall for the Other and Vehicle categories are much better than the VRU category. VRU classification is more difficult as VRU is often covered by few angular bins as compared to the other categories.
\begin{table}[hbt!]
\begin{tabular}{cccc}
\hline
Relative gap & Absolute gap & Success rate  & Smoothness \\ \hline
44.14 & 1.95 & 77.59 & 0.77 \\
\hline
\\
\hline
\end{tabular}
\quad
\begin{tabular}{cccc}
\hline
& Vehicle & VRU & Other\\
\hline
Precision  & 0.92 & 0.73  & 0.98\\
\hline
Recall & 0.92 & 0.66 & 0.98\\
\hline
\end{tabular}
\centering
\caption{3D freespace regression metrics (left) and classification metrics (right).}
\label{table:freespace_result_solo}
\end{table}
\begin{table}[t]
	\begin{tabular}{llcccc}
	\hline
	Radial (meters) & Angular (degrees)  & Success Rate (\%) & Absolute gap (m)\\
	\hline
	0-10  &  [-45, +45] & 86.43  & 0.91\\
        10-20   &  [-45, +45] & 83.02 & 1.40\\
        20-30   &  [-45, +45] & 73.74 & 2.70\\
        30-50   &  [-45, +45] & 64.52 & 4.72\\
        50-80   &  [-45, +45] & 57.05 & 8.21\\
        80-120  &  [-45, +45] & 42.39 & 18.96\\
        120-200 &  [-45, +45] & 1.81 & 54.38\\
	\hline
        0-10  &  [-135, +135] & 82.68  & 0.93\\
        10-20   & [-135, +135] & 76.54 & 1.67\\
	  20-30   &  [-135, +135] & 67.85 & 3.11\\
        30-50   &  [-135, +135] & 58.96 & 5.31\\
        50-80   &  [-135, +135] & 51.46 & 9.31\\
        80-120   &  [-135, +135] & 38.84 & 15.39\\
        120-200   &  [-135, +135] & 0.04 & 61.2\\
	\hline
	\end{tabular}
	\centering
	\caption{Freespace regression metrics for different radial ranges and field-of-views.}
	\label{tab:freespace_result_sector}
\end{table}
\subsubsection{3D Parking space}
\begin{table}
    \begin{tabular}{c|ccc}
    \hline
    & AP & mean IoU & F1\\
    \hline
    Angled & 0.68 & 0.86 & 0.75 \\
    Parallel & 0.19 & 0.82 & 0.37 \\
    Perpendicular & 0.57 & 0.85 & 0.67 \\ 
    All & 0.58 & 0.85 & 0.68\\
    \hline
    \end{tabular}
    \centering
    \caption{Parking space detection performances.}
    \label{table:parking_results_solo}
\end{table}
Displayed in Table \ref{table:parking_results_solo}, the parking space detection outcomes are presented. It's noteworthy that the mean IoU for true positive detections hovers around 86\%, suggesting a strong alignment between the majority of detections and the actual ground truth labels.
However, it's observed that the parallel parking space category exhibits the lowest performance. This can be attributed to labeling intricacies. The dimensions of parallel parking spaces exhibit significant variability, contributing to difficulties in accurate labeling. Moreover, many curbside parking spaces lack precise width regulations, compelling labelers to rely on subjective judgments for width determination. Consequently, the labeling process for such spaces becomes inherently inconsistent.
%Table \ref{table:parking_results_solo} reports parking space detection results. We see that the mean IoU of all the detections that were counted as true positives is close to 86\%, which indicates that the majority of the detections highly overlap with the ground truth labels. We also observe that the lowest performing class is the parallel parking space, which can be attributed to challenges in labeling these parking spaces. The dimensions of parallel parking spaces are highly variable. In addition, many parking spaces on the side of the street do not have strictly enforced widths. As such, labelers have to rely on their judgment to determine how wide each parking space should be, which in turn makes the labeling of such spaces highly inconsistent.
\subsection{Single task vs Multi-task Learning}
For this experiment, we establish distinct $c_t$ parameters (as introduced in Section \ref{sec:loss_balancing}) specifically as $[5,3,1]$ for the individual obstacle, parking space, and freespace tasks. The rationale behind this selection is to emphasize obstacle detection by assigning a higher weight, while relatively less emphasis is placed on the freespace task due to its relatively lower complexity.
Performance comparison between the multi-task model and single task models is detailed in Table \ref{table:single_vs_multi_tasks}. It's observed that obstacle and parking detection accuracies hold up comparably to those achieved by the single task models. Although the freespace task experiences a 9.5\% drop, its accuracy remains commendable. This outcome underscores the substantial efficacy of our proposed multi-task loss balancing algorithm.

%In this experiment, we set the $c_t$ parameters (defined in Sec. \ref{sec:loss_balancing}) be $[5,3,1]$ for obstacle, parking space and freespace tasks separately. 
%Intuitively, we want the network pay more attention to the obstacle detection task, and less to the freespace task as obstacle detection is much more difficult than others. Table \ref{table:single_vs_multi_tasks} compares the performances of the multi-task model against the single task models. We observe that obstacle and parking detection accuracy are comparable to those of single task models. The freespace task drops $9.5\%$, but its accuracy still remains high. This result indicates the impressive effectiveness of our proposed multi-task loss balancing algorithm.
\begin{table}[hbt!]
    \begin{tabular}{c|c|c|c}
    \hline
     & Obstacle (mAP) & Freespace (Success rate) & Parking space (F1) \\ \hline
    Single task & 0.46 & 77.59 & 0.68  \\
    Multi-task & 0.46 & 70.19 & 0.68\\ 
    \hline
    \end{tabular}
    \centering
    \caption{Single task vs. multi-task.}
    \label{table:single_vs_multi_tasks}
\end{table}
\subsection{IPM vs MLP based 2D-to-BEV View Transformation}
This study aims to showcase the advantages of employing a learning-based technique over an IPM method to convert 2D image features to BEV features. For this experiment, we focus on the obstacle detection task. The outcomes are summarized in Table \ref{tab:ipm_vs_mlp}, where our proposed 2D-to-BEV approach demonstrates a remarkable performance improvement over the traditional IPM method.
\begin{table}[hbt!]
	\begin{tabular}{lccccc}
	\hline
	 & Vehicle & Truck & Person  & Bike-rider & \\
	 Method & AP & AP & AP  & AP & mAP\\
	\hline
	IPM & 0.49 & 0.32 & 0.27 & 0.33 & 0.30 \\
	MLP & 0.63 & 0.38 & 0.35 & 0.48 & 0.46 \\
	\hline
	\end{tabular}
	\centering
	\caption{IPM versus MLP based 2D-to-BEV view transformation for obstacle detection.}
	\label{tab:ipm_vs_mlp}
\end{table} 

\subsection{Generalization to Different Vehicle Lines}
\begin{figure}
\centering
\begin{subfigure}[b]{.2\textwidth}
    \includegraphics[width=1.0\textwidth]{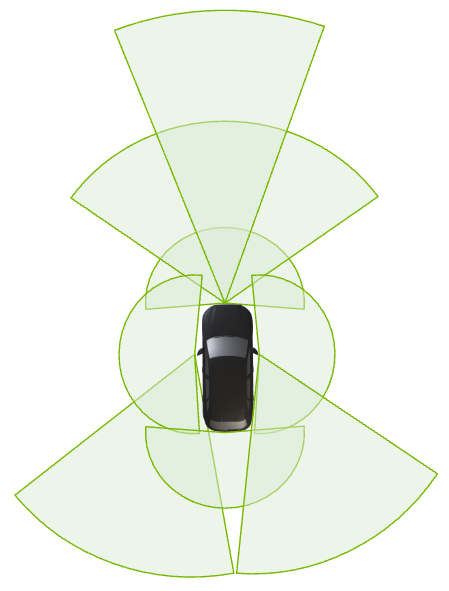}
    \caption{Car platform.}
\end{subfigure}
\begin{subfigure}[b]{.27\textwidth}
    \includegraphics[width=1.0\textwidth]{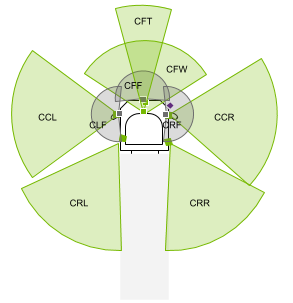}
    \caption{Truck platform.}
\end{subfigure}
\caption{Different camera sensor setups for cars and trucks.}
\label{fig:cam_configs}  
\end{figure}

In order to validate the robustness and scalability of our proposed architecture, we assess the performance of NVAutoNet on a truck platform, despite its initial development for a car platform. While there exist minor disparities in intrinsic parameters between the two platforms, the variations in extrinsic parameters are notably significant (see Figure \ref{fig:cam_configs}). Nevertheless, we demonstrate that deploying our perception model on a distinct platform doesn't necessitate extensive model redesign or extensive data collection. Remarkably, achieving satisfactory results only requires fine-tuning the model using a limited training dataset.
\begin{table}
    \begin{tabular}{*{8}{c}}
    \hline
    & Pre-trained & Trained/Finetuned & \multicolumn{5}{c}{Dataset size (Truck Platform)}  \\ \hline
    &  &  & 50K & 75K & 100K & 125K & 150K \\ \hline
    Model-A & No & Yes & 0.146 & 0.180 & 0.192 & 0.209 & 0.210 \\
    Model-B & Yes & No & 0.168 & 0.168 & 0.168 & 0.168 &0.168  \\
    Model-C & Yes & Yes &0.286 & 0.292 & 0.294 & 0.294 & 0.300  \\ \hline
    \end{tabular}
    \centering
    \caption{Transfer learning experiments for NVAutoNet. mAP scores for obstacle detection are reported. Model-A trained from scratch using different dataset sizes collected from the truck platform; Model-B previously trained using data collected from \emph{\textbf{the car platform}}; Model-C fine-tuned from model-B using data collected from the truck platform.}
    \label{table:truck_dataset_size}
\end{table}

\textbf{Datasets}: This study focuses on the 3D obstacle detection task. We gathered data from both car and truck platforms for training purposes. Specifically, we collected 850K scenes for car model training (car-dataset-train) and 150K scenes for truck model training (truck-dataset-train), along with 26K scenes for truck model validation (truck-dataset-val). The label distribution is roughly similar between the two platforms. To explore the effect of training set size, we created smaller subsets from the 150K truck-dataset-train, ranging from 50K to 150K scenes.

Our comparison involves three models: Model-A trained solely with truck-dataset-train, Model-B pretrained with car-dataset-train, and Model-C fine-tuned from Model-B using truck-dataset-train. All models are tested on truck-dataset-val, and the results are presented in Table \ref{table:truck_dataset_size}. As anticipated, Model-A's performance improves as the dataset size increases. Interestingly, even with only 50K scenes, Model-B, which has never seen data from a truck platform, outperforms Model-A. This illustrates the network's generalization capability. Notably, Model-C outperforms both Model-A and Model-B by a significant margin. It's worth noting that Model-C's performance improvement with larger fine-tuning datasets is not as substantial as Model-A's. This suggests that only a small amount of data is required for fine-tuning when deploying the NVAutoNet model, previously trained for one platform, onto another. These findings reaffirm the robustness and scalability of our proposed NVAutoNet, crucial qualities for practical production applications.

\section{Conclusion and Future Work}
NVAutoNet emerges as a specialized perception network designed to meet the unique demands of autonomous vehicles. Unlike many existing 3D perception methods, it strikes a balance between accuracy and computational efficiency, onboard chip compatibility, and adaptability to diverse vehicle types. NVAutoNet processes synchronized camera images to predict 3D signals such as obstacles, freespaces, and parking spaces. Its architecture employs efficient convolutional networks for image and BEV backbones, optimized for efficiency with TensorRT. The image-to-BEV transformation uses simple linear layers and BEV look-up tables for rapid yet accurate inference.

Trained extensively on proprietary data, NVAutoNet consistently attains high perception accuracy while maintaining real-time performance, achieving 53 frames per second on the NVIDIA DRIVE Orin SoC. It excels in handling sensor mounting deviations across car models and adapts seamlessly to various vehicle types through streamlined model fine-tuning procedures. NVAutoNet stands as a significant advancement in autonomous vehicle perception, addressing key challenges and offering efficient and reliable 3D perception for real-world scenarios.

Extending BEV perception to truly 3D perception (e.g., 3D volumetric occupancy perception) will enable higher levels of autonomy such as L4/L5 self-driving. But that transition is seen to be very challenging due to its high memory and computation consumption. Far range perception (e.g., up to 300 meters) will be necessary to increase driving safety and comfortableness. Holistic scene understanding, where not only dynamic objects but also static objects such as lane graphs, and relations/associations between them are predicted, becomes more relevant to unlock autonomous mapless driving.

\section*{Acknowledgments}
We would like to thank the NVIDIA Maglev and ML foundation teams for the machine learning infrastructure support, thank the NAS team for the network optimization, thank Tero Kuosmanen for the initial idea of the loss balancing algorithm and special thank to the whole DRIVE AV team for data collection, network deployment, iteration and on-road testing.
%We would like to thank the NVIDIA Maglev and ML foundation teams for the machine learning infrastructure support, thank the NAS (neural architectural search) team for the network optimization, thank Tero Kuosmanen for the initial idea of the loss balancing algorithm and special thank to the whole DRIVEAV team for data collection, network deployment, iteration and on-road testing. 

%Bibliography
\bibliographystyle{unsrt}  
\bibliography{references}

\end{document}